\documentclass{article}
\usepackage{ijcai21}

\usepackage{times}
\usepackage{soul}
\usepackage{url}
\usepackage[hidelinks]{hyperref}
\usepackage[utf8]{inputenc}
\usepackage[small]{caption}
\usepackage{graphicx}
\usepackage{amsmath}
\usepackage{amsthm}
\usepackage{booktabs}
\usepackage{bm}
\usepackage{multirow}
\usepackage[ruled,vlined]{algorithm2e}
\usepackage[noend]{algpseudocode}

\title{Adv-Makeup: A New Imperceptible and Transferable Attack on Face Recognition}

\author{
Bangjie Yin$^{1*}$
\and
Wenxuan Wang$^{2*}$\and
Taiping Yao$^{1*\dagger}$\and
Junfeng Guo$^3$ \and
Zelun Kong$^3$ \and \\
Shouhong Ding$^{1\dagger}$\and
Jilin Li$^1$ \And
Cong Liu$^3$ \\
\affiliations
$^1$Youtu Lab, Tencent, Shanghai, China,\\
$^2$Fudan University, Shanghai, China,\\
$^3$The University of Texas at Dallas, Dallas, Texas, USA \\
\emails
\{jamesyin10,  wxwang.iris, gjf199509236817\}@gmail.com,
\{taipingyao,  ericshding, jerolinli\}@tencent.com,
\{Zelun.Kong, cong\}@utdallas.edu
}

\begin{document}

\maketitle

\renewcommand{\thefootnote}%
{\fnsymbol{footnote}}
\footnotetext[1]{indicates equal contributions.} 
\footnotetext[2]{indicates corresponding author.}

\begin{abstract}
Deep neural networks, particularly face recognition models, have been shown to be vulnerable to both digital and physical adversarial examples. However, existing adversarial examples against face recognition systems either lack transferability to black-box models, or fail to be implemented in practice. 
In this paper, we propose a unified adversarial face generation method - Adv-Makeup, which can realize imperceptible and transferable attack under black-box setting. 
Adv-Makeup develops a task-driven makeup generation method with the blending module to synthesize imperceptible eye shadow over the orbital region on faces. And to achieve transferability, Adv-Makeup implements a fine-grained meta-learning adversarial attack strategy to learn more general attack features from various models. 
Compared to existing techniques, sufficient visualization results demonstrate that Adv-Makeup is capable to generate much more imperceptible attacks under both digital and physical scenarios. Meanwhile, extensive quantitative experiments show that Adv-Makeup can significantly improve the attack success rate under black-box setting, even attacking commercial systems.
\end{abstract}

\section{Introduction}

Deep neural networks (DNNs) are shown to achieve state-of-the-art, and even human-competitive performance in many application domains, such as face recognition.
However, recent works have proved that DNNs are vulnerable to adversarial examples \cite{goodfellow2014explaining,dong2018boosting,papernot2017practical,liu2019universal}, which are generated by adding perturbations over clean samples. 
These adversarial examples ~\cite{deng2019arcface,dong2019efficient} are effective against state-of-the-art face recognition (FR) systems, which are pervasively seen in a wide range of applications, from airport security check to mobile phone unlock payment. 

\begin{figure}
\begin{centering}
\includegraphics[scale=0.4]{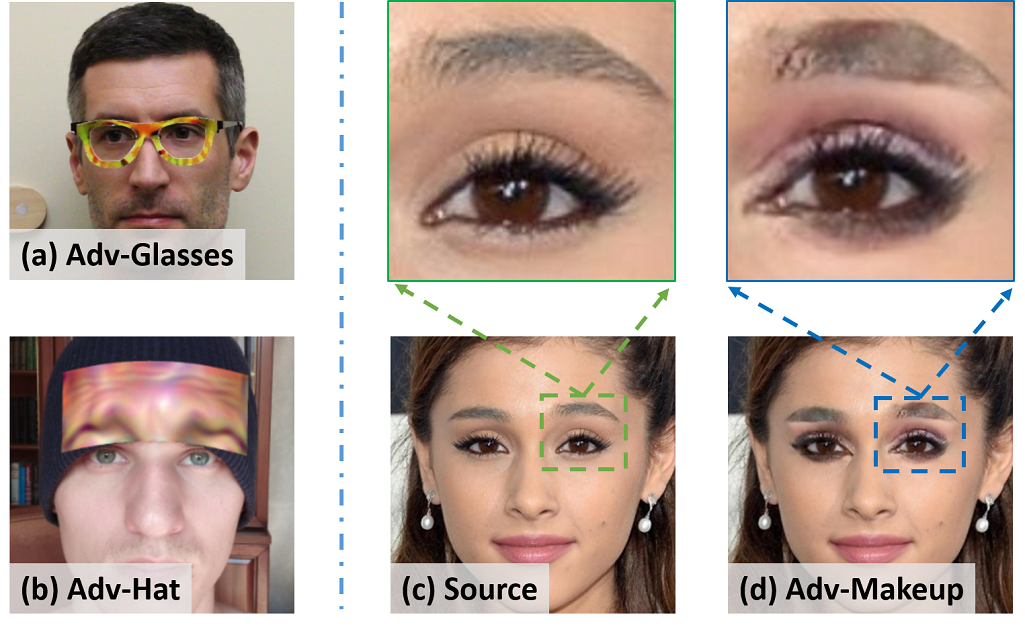} 
\par\end{centering}
\caption{Illustration of the adversarial faces generated by Adv-Glasses, Adv-Hat, and our Adv-Makeup. Impersonation attacks generated by Adv-Glasses and Adv-Hat are shown on the left. Source face, Adv-Makeup attack, and the corresponding zooming-in images are illustrated on the right. \label{fig:intro}}
\end{figure}

Unfortunately, existing adversarial example generation methods on FR models exhibit several critical limitations that may severely constraint their applicability in practice. 
Specifically, we highlight the following limitations:
(1) \emph{Most existing methods are impractical and ineffective under physical scenarios}.  A rich set of methods  \cite{dong2019efficient,zhong2020towards,song2018attacks,deb2019advfaces} could generate inconspicuous and strong perturbations over the entire facial images, focusing on the digital-world setting. Whereas, such global perturbations cannot be applied in realistic applications, because the attackers can only be able to control their (physical) appearance, for example, it is impossible to add noise in the background area.
(2) \emph{Recent adversarial methods ~\cite{sharif2016accessorize,komkov2019advhat,brown2017adversarial} which generate physical-implementable samples are rather noticeable which could be easily detected.} As illustrated in Fig.\ref{fig:intro} (a) and (b), the adversarial samples generated under such methods are visually obvious noticeable by human eyes, such as a pair of eyeglass, or imitating sticker located on the hat.
(3) \emph{The attacks are not transferable, \emph{i.e.}, not applicable to black-box FR models.} Existing methods mostly use the gradients of target model to construct adversarial examples. However, the target model is often inaccessible directly in practical scenarios. Thus, these attacks are very difficult to successfully fool the black-box FR model.

To tackle the above-discussed challenges simultaneously, we propose a novel imperceptible and transferable attack technique, Adv-Makeup, for face recognition. 
Firstly, the generated attacks shall be feasible in real-world applications. 
Adv-Makeup targets at a common and practically implementable scenario: adding makeup to eye regions which shall mislead FR models yet being visually unnoticeable (\emph{i.e.}, appearing as natural makeup). 
We develop a makeup generation module to synthesize realistic eye shadow on the source identity. 
To achieve imperceptible attack, we also apply a makeup blending method to alleviate the style and content differences between the source faces and generated eye shadows,
which can further improve the naturalness of the resulting adversarial faces.
As shown in Fig.\ref{fig:intro} (c) and (d), the synthesized faces of Adv-Makeup with generated eye shadow appear natural and inconspicuous.
Moreover, to enhance attack transferability, the adversarial examples should be model-agnostic. We thus introduce a fine-grained meta-learning attack strategy to learn better-generalized adversarial perturbations to improve the transferability of generated attack faces under the black-box scenario.


Our contributions are summarized as follows, 
(1) We propose a novel general Adv-Makeup technique, which can implement an imperceptible and transferable attack against face recognition models under both digital and physical scenarios.
(2) To enhance the imperceptibility of attacks, we introduce a makeup generation module, which can add natural eye shadow over the orbital region. Moreover, a makeup blending strategy is proposed to ensure the consistency of style and content between source and generation, which can further improve the fidelity of synthetic faces.
(3) We also propose a task-driven fine-grained meta-learning adversarial attack strategy to guarantee the attacking ability of the generated makeup, especially improving the transferability of adversarial examples on black-box victim models.
(4) The comprehensively designed experiments over the makeup dataset and LFW dataset show that Adv-Makeup can generate imperceptible and transferable adversarial examples. A case study implementing Adv-Makeup in the physical world against two popular commercial FR platforms also demonstrates its efficacy in practice.

\section{Related Work}

\textbf{Adversarial Attacks. }
Many attack algorithms have recently been proposed to show the deep learning models are broadly vulnerable to adversarial samples. 
For the white-box attack, there are
lots of gradient-based approaches, such as \cite{goodfellow2014explaining,carlini2017towards,madry2017towards,dong2018boosting}.
However, such white-box attacking requires the adversarial examples to have full access to target models, which is unrealistic in real applications. 
Under a more practical scenario, there exist many black-box attack works \cite{liu2019universal,papernot2017practical}, where the adversary can not access the target model except only the outputs can be fetched. 
Whereas, the transferability of adversarial examples is the main challenge of the task.
Even there already exist many studies trying to improve the transferability
\cite{dong2019evading,xie2019improving}, the attack success rates of black-box attacks still decrease with a large margin. 
Moreover, the above attack methods apply perturbations to the entire input, which is impractical in the real world.
In contrast, our Adv-Makeup method focuses on achieving implementable physical attack with high transferability.

\begin{figure*}
\begin{centering}
\includegraphics[scale=0.39]{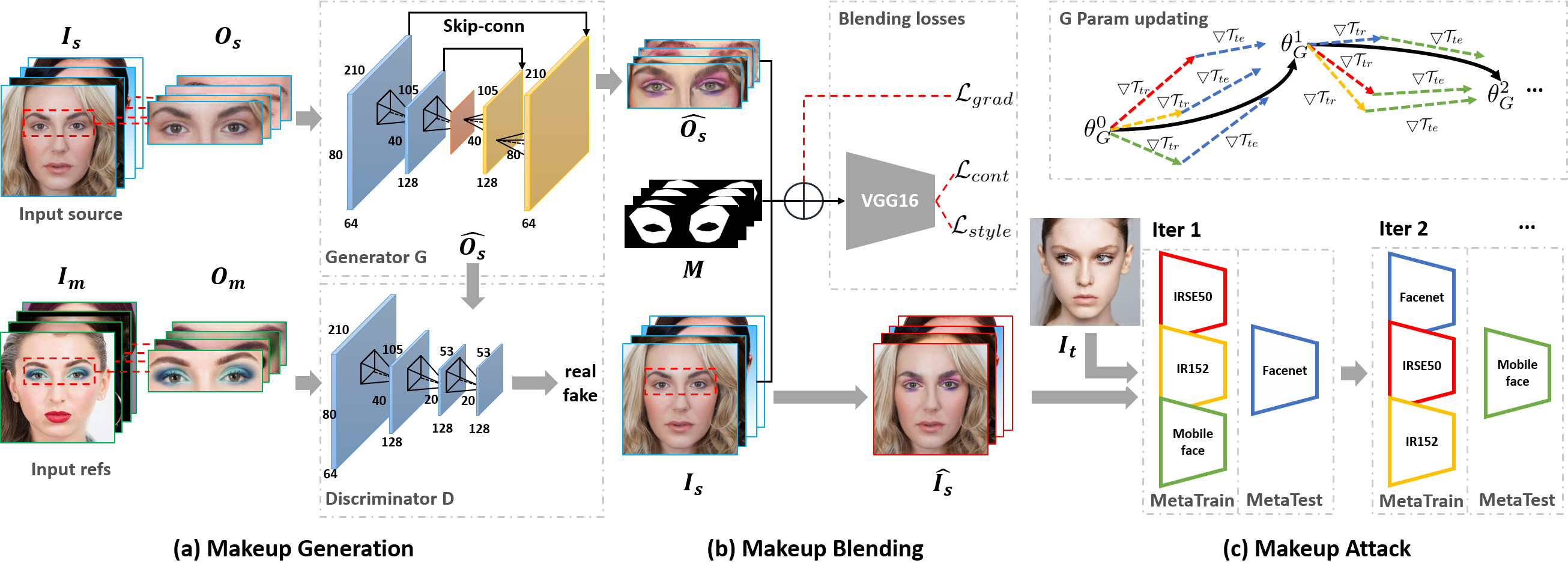} 
\par\end{centering}
\caption{
Overview of unified Adv-Makeup framework, which contains makeup generation, makeup blending, and fine-grained meta-learning adversarial attack. (a) Makeup generation generates faces with eye shadow over the orbital regions. (b) Makeup blending is further applied to make the eye shadow more naturally to achieve imperceptible generation. (c) Fine-grained meta-learning adversarial attack realize the transferable impersonation attack. Different colors in `G param updating' denote various victim face recognition models.
\label{fig:framework}}
\end{figure*}

\noindent \textbf{Adversarial Attacks on Face Recognition. }
Many studies have been proposed to evaluate the robustness of advanced FR models. 
In general, there are two kinds of attack, \emph{i.e.}, untargeted (dodging) and targeted (impersonation) attacks. Different from the dodging attack, which only aims to lower the similarity confidence of the same-identity pairs, the impersonation attack tries to improve the feature similarity between the target and source images of different identities.
Such a technical difference makes impersonation attack much harder than dodging attack. Therefore, we mainly focus on the impersonation attack.
As for the digital facial attacks, \cite{zhu2019generating} firstly attempt to transfer eye-makeup to perform attacks on face recognition, and \cite{dong2019efficient,zhong2020towards,sharif2019general,song2018attacks,deb2019advfaces},
however their adversarial examples either have obvious artifacts or can not directly apply in the physical world.
Considering the wide application of FR in real-world applications, studies on physical attacks attract lots of researchers.
\cite{sharif2016accessorize} proposes a patch-based attack using optimization-based methods to add perturbations to the eyeglass region. 
Such attacks over the wearing hat are also generated in \cite{komkov2019advhat}.
However, these adversarial patches are perceptible to humans, and are mostly applied in the white-box setting, which is impractical in real-world scenarios. 
Different from prior works, our Adv-Makeup method can boost the similarity between the adversarial examples and targets under black-box scenario in both digital and physical conditions, and the synthesized makeup is imperceptible to be noticed.

\section{Methodology}

\subsection{Overview}
To generate imperceptible and transferable adversarial examples for the FR models, we propose a unified framework, \textit{Adv-Makeup}, which can generate makeup-like perturbations over the orbital region of two eyes and learn better-generalized adversarial features from various models.
As shown in Fig.~\ref{fig:framework}, the Adv-Makeup framework comprises of three components, \emph{i.e.}, makeup generation, makeup blending, and makeup attack. 
Taking the non-makeup source facial image and random face with cosmetics as inputs, the makeup generation can synthesize the face with realistic eye shadow.
To further improve the visually-indistinguishable quality of generated adversarial samples, the makeup blending module helps to fuse the generated orbital region to the source face.
Meanwhile, a fine-grained meta-learning adversarial attack module is introduced to make the generated faces adversarial as well as further improve the transferability of the attacks under black-box condition.

\subsection{Makeup Generation}

The goal of Adv-Makeup is to generate inconspicuous adversarial faces. Considering that eye-makeup is the most common facial alterations in daily life, owns various appearances, and can be implemented easily, besides, the eye area is a distinguishable region when performing face recognition, thus, we intend to firstly apply a makeup generation module to synthesize natural eye shadow over the orbital regions.
To achieve this goal, we naturally turn to generative adversarial networks (GANs), which are widely used to synthesize images \cite{zhai2019lifelong}
and conduct makeup transfer \cite{wang2020fm2u,zhu2019generating}.
As shown in Fig.~\ref{fig:framework},
the makeup generation contains a generator $G$ to synthesize eye shadow, and a discriminator $D$ to encourage perceptual realism of the generated images.

With the input of source image $I_{s}\in D_{s}$ and makeup image $I_{m}\in D_{m}$, firstly we apply face alignment method \cite{zhang2016joint} to generate corresponding landmarks for different faces. We align the faces according to affine matrix, which will also be used to align the landmarks. The landmarks surrounding two eyes are used for generating the 0-1 mask, and get the orbital regions $O_{s}$ and $O_{m}$. 
Then the proposed generator takes orbital region $O_{s}$ of the source image as input, and output $\hat{O_{s}}=G(O_{s})$ with generative eye shadow. 
Next, the resulting makeup orbital region can be calculated through 
$ O_{s} \odot (1-M) + \hat{O_{s}} \odot M $, where $\odot$ is the element-wise product, $M$ with equal-size of the $ O_{s} $ is a binary mask indicating the orbital region over the face. This region will be attached to the source face $I_{s}$ to get the resulting face $\hat{I_{s}}$.


To improve the quality of the generation, we also introduce a discriminator encouraging the generated images to be more perceptually natural. 
The discriminator takes the real cosmetic orbit $O_{m}$ of the makeup image and the output $\hat{O_{s}}$ with synthesized eye shadow from generator as input, 
and it enforces the generation as natural as real eye shadow. 
The two-player `mini-max' game  makes the generation visually natural, so we define the generator loss $\mathcal{L}_{gen}$ and discriminator loss $\mathcal{L}_{dis}$ as, 

\begin{center}
\vspace{-4mm}
\begin{equation}
\mathcal{L}_{gen}=E_{O_{s}}[log(1-D(G(O_{s})))]
\label{eq: gen}
\end{equation}
\par\end{center}

\begin{center}
\vspace{-4mm}
\begin{equation}
\begin{split}
\mathcal{L}_{dis}=-[E_{O_{m}}[log(D(O_{m}))]+E_{O_{s}}[log(1-D(G(O_{s})))]]
\label{eq: dis}
\end{split}
\end{equation}
\par\end{center}

\subsection{Makeup Blending}

By applying such makeup generation, we can generate natural eye shadow over the orbital region, however, directly attaching the synthesized orbit region to the source image by using mask would yield obvious style differences in content and abrupt intensity changes at the boundary. 
To eliminate the noticeable artifacts at the boundary and the style induced by the eye-shadow patches, we propose a makeup blending method to realize the imperceptible generation.

First, a gradient domain constraint is proposed, which we convert to a differentiable loss function, to alleviate the changes over generated boundary.
By minimizing the loss function, the image details of generated face $\hat{I_{s}}$ are able to be preserved while shifting the color to match the original image $I_{s}$. 
The gradient constraint loss $\mathcal{L}_{grad}$ is defined as,
\begin{center}
\vspace{-4mm}
\begin{equation}
\mathcal{L}_{grad}=\parallel[\bigtriangledown I_{s}\odot (1-M^{*}) + \bigtriangledown h(\hat{O_{s}}) \odot M^{*}] - \bigtriangledown \hat{I_{s}}\parallel_{2}^{2} 
\label{eq: poisson}
\end{equation}
\par\end{center}
where $\bigtriangledown$ means the gradient operator, and $ M^{*} $ is a 0-1 mask expanded from $ M $ with the same size as $ I_{s} $, $h$ attach $ \hat{O_{s}} $ to a all $ 0 $s matrix with the size of $ I_{s} $.

Except for applying the $\mathcal{L}_{grad}$ to control the smoothness and style over the $ \hat{I_{s}} $, we also try to enhance the integration of style and content to better advance the naturalness of synthesized eye-shadow.
Therefore, inspired by the \cite{gatys2016image}, we utilize a pre-trained VGG16 model to calculate the style and content loss functions, \emph{i.e.}, $ \mathcal{L}_{style} $ and $\mathcal{L}_{cont}$, pushing the style of eye-shadow patches closer to the source image and also preserving the content of the synthesized region. The content loss and style loss can be defined as,

\begin{center}
\vspace{-4mm}
\begin{equation}
\mathcal{L}_{cont}=\sum_{p=1}^{P}\frac{\alpha _{p}}{2N_{p}M_{p}}\sum_{j=1}^{N_{p}}\sum_{k=1}^{M_{p}}[(A_{p}[\hat{I_{s}}]-A_{p}[I_{s}])\odot M^{*}]_{jk}^{2}
\label{eq: content}
\end{equation}
\par\end{center}

\begin{center}
\vspace{-4mm}
\begin{equation}
\mathcal{L}_{style}=\sum_{p=1}^{P}\frac{\beta_{p}}{2N_{p}^{2}}\sum_{j=1}^{N_{p}}\sum_{k=1}^{N_{p}}(B_{p}[\hat{I_{s}}]-B_{p}[I_{s}])_{jk}^{2}
\label{eq: style}
\end{equation}
\par\end{center}
where P is the number of convolutional VGG layers, $N_{p}$ is the number of channels in activation, $M_{p}$ is the number of flattened activation values in each channel.
$A_{p}[\cdot]\in R^{N_{p}\times M_{p}}$ is an activation matrix computed from a deep network at corresponding layers. 
$B_{p}[\cdot]=A_{p}[\cdot ]A_{p}[\cdot ]^{T}\in R^{N_{p}\times N_{p}}$ is the Gram matrix of features extracted at a set of VGG layers, and the $\alpha_{p}$ and $\beta_{p}$ are the weights to balance the contribution of each layer when calculating content and style loss.

\subsection{Makeup Attack}
{\bf Adversarial Attack.} To make the generated makeup adversarial, we introduce an attack loss function against FR models. 
Leveraging a pre-trained face feature extractor $F$, the loss of impersonating attack ($ \mathcal{L}_{target} $) can be expressed as,

\begin{center}
\vspace{-4mm}
\begin{equation}
\mathcal{T}=1-cos[F(I_{t}),F(\hat{I_{s}})]
\end{equation}
\par\end{center}
where $ I_{t} \in D_t$, denoting the target image. The loss objective is utilized to make the source image wearing disturbed makeup $ \hat{I_{s} }$ to be close to the target image $ I_{t} $.

Nevertheless, for the black-box attack, such adversarial examples can not `generalize' well on gradient-unknown models, resulting in low transferability. 
Recently, through finding the common-shared gradient direction among different victim models, ensemble training \cite{liu2016delving,dong2018boosting} can significantly improve the transferability of adversarial perturbations. 
In spite of that, ensemble training still has high chance to `overfit' to the white-box models, thus perform ineffectively against the unknown victim models.  
Meta-learning has been applied in many research fields, like domain generation, and there are also some relevant applications in adversarial attacking/defence \cite{du2019query,yin2018adversarial}. However, they either focus on query-based black-box attack or enhancing network robustness. Inspired by \cite{shao2020metaantisppof}, we think meta-learning can also push the frontier of transferable attacks, therefore the fine-grained meta-learning adversarial attack is proposed to boost the black-box transferability.

\noindent{\bf Meta-Train.} Assume we have $L$ pre-trained victim FR models, $ \mathcal{F}={F_1, F_2, ..., F_L} $, then we choose $L-1$ models to be meta-train models with performing impersonating attacks in each model,

\begin{center}
\vspace{-4mm}
\begin{equation}
\mathcal{T}_{tr}^{i}(\theta_{G})= 1-cos[F_i(I_{t}),F_i(G(I_{s}))]
\end{equation}
\par\end{center}
where $ i \in \left\{ 1,...,L-1\right\} $ and $ \theta_{G} $ is the parameter of generator, $G(I_{s})$ indicates the faces with generated eye shadow from the makeup generator accompanying with makeup blending method. $\mathcal{T}_{tr}^{i}(\theta_{G})$ indicates that different models share a same generator, collecting the useful gradient information from each of the $ L-1 $ models to be a proper prior. 
To obtain $ i_{th} $ copy when updating $ \theta_{G} $, we have $\theta_{G}^{'i} \leftarrow \theta_{G} - \alpha_1 \bigtriangledown_{\theta_{G}} \mathcal{T}_{tr}^{i}(\theta_{G})$.

\noindent{\bf Meta-Test.} Moreover, we still have one remaining victim model, $F_L$, which will be the meta-test model. After obtaining the updated parameters, we conduct the impersonating attack by using the generator with $ \theta_{G}^{'i} $ on $F_L$,

\begin{center}
\vspace{-4mm}
\begin{equation}
\mathcal{T}_{te}^{i}(\theta_{G}^{'i})= 1-cos[F_L(I_{t}),F_L(G(I_{s}))]
\end{equation}
\par\end{center}

\noindent{\bf Meta-Optimization.} To collect all the information in two stages, we propose a joint optimization strategy. The updating process can be written as Eq. \ref{eq: metaopt}, where $\alpha_1$ and $\alpha_2$ are the balancing weights for generation and attacking, and
$\beta_1$, $\beta_2$ and $\beta_3$ are the
corresponding parameters of makeup blending. Overall, the whole training process is illustrated in Alg.\ref{alg:algorithm}.

\begin{center}
\vspace{-4mm}
\begin{equation}
\begin{split}
\theta_{G}^{''} \leftarrow \theta_{G} - \alpha_1 \sum_{i=1}^{L-1}
\bigtriangledown_{\theta_{G}} (\mathcal{T}_{tr}^{i}(\theta_{G}) + \mathcal{T}_{te}^{i}(\theta_{G}^{'i})) - \\
\alpha_2 \bigtriangledown_{\theta_{G}} \mathcal{L}_{gen}(\theta_{G}) - 
\beta_1 \bigtriangledown_{\theta_{G}} \mathcal{L}_{grad}(\theta_{G}) - \\
\beta_2 \bigtriangledown_{\theta_{G}} \mathcal{L}_{cont}(\theta_{G}) - 
\beta_3 \bigtriangledown_{\theta_{G}} \mathcal{L}_{style}(\theta_{G})
\end{split}
\label{eq: metaopt}
\end{equation}
\par\end{center}

\begin{algorithm}[t!]
  \caption{The proposed Adv-Makeup.}
  \label{alg:algorithm}
  \begin{algorithmic}
    \Require \\
      \hspace*{0.02in} {\bf Input:}
      Source image $ I_s \in D_s$;
      makeup image $ I_m \in D_m$;
      target image $ I_t \in D_t$;
      pre-trained face models $ \mathcal{F}={F_1, F_2, ..., F_L} $;
      iterations $ T $;
      generator $ G $ and discriminator $ D $.\\
      \hspace*{0.02in} {\bf Initialization:} 
      Model parameters $ \theta_G $, $ \theta_D $;
      hyperparameters $\alpha_{1},\alpha_{2},\beta_{1},\beta_{2},\beta_{3}$.
  \end{algorithmic}
  \begin{algorithmic}[1]
    \Ensure
      Model parameters $ \theta_G^{*} $, $ \theta_D^{*} $.
    \State {\bf for} each $ i\in T $ {\bf do}
    \State    Update D: Compute $\mathcal{L}_{dis}$, then update $ \theta_{D}^{'} \leftarrow \theta_{D} - \bigtriangledown_{\theta_{D}} \mathcal{L}_{dis}(\theta_{D}) $
    \State Update G:
    
    \State \quad {\bf Meta-train:} \\
     \qquad Randomly select $ L-1 $ models from $ \mathcal{F} $ as meta-train models\\
     \qquad Calculate ($ \mathcal{T}_{tr}^{i} $ , $ \mathcal{L}_{gen} $, $ \mathcal{L}_{grad} $, $ \mathcal{L}_{cont} $, $ \mathcal{L}_{style} $) and then update $ \theta_{G}^{'i} \leftarrow \theta_{G} - \alpha_{1} \bigtriangledown_{\theta_{G}} \mathcal{T}_{tr}^{i}(\theta_{G}) $
    \State \quad {\bf Meta-test:}\\
     \qquad Use the remaining $ F_L $ to be meta-test model\\
     \qquad Calculate $ \mathcal{T}_{te}^{i} $
    \State \quad {\bf Meta-optimization:}
    \State \qquad With Eq. \ref{eq: metaopt}
    \State {\bf endfor}
    \State $ \theta_G^{*} = \theta_{G}^{''}$, $ \theta_D^{*} =  \theta_{D}^{'}$\\
    \Return $ \theta_G^{*} $, $ \theta_D^{*} $;
  \end{algorithmic}
\end{algorithm}

\section{Experiment}
\subsection{Experimental Setup}

\begin{figure*}
\begin{centering}
\includegraphics[scale=0.50]{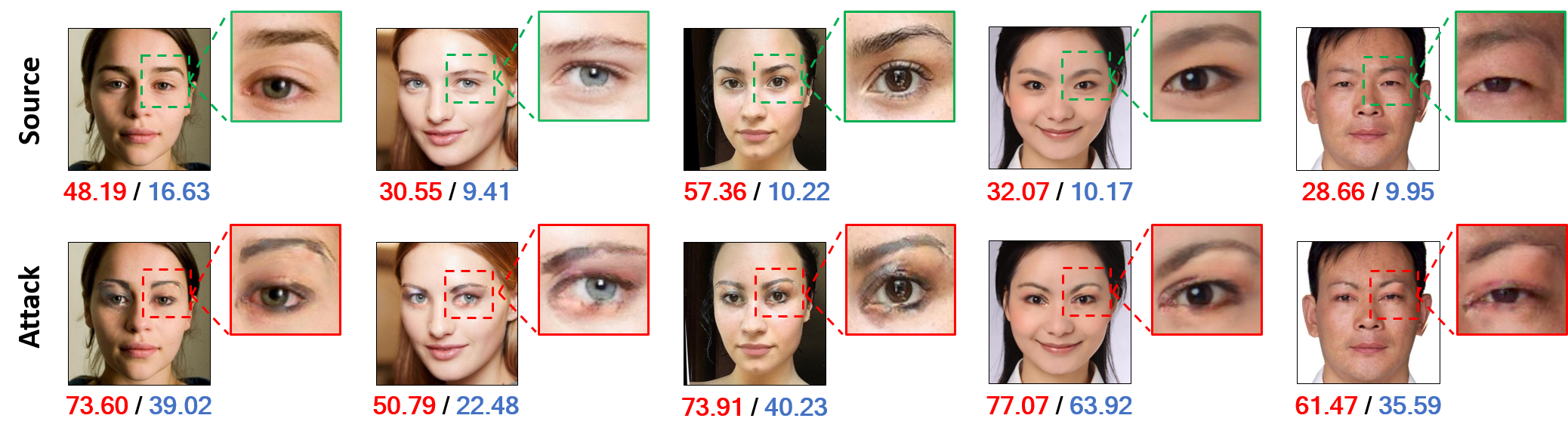}
\par\end{centering}
\caption{Visualization of generated adversarial faces. There exist various illumination, pose, gender, and makeup intensity changes. The target identity is the same target female in the first row of Fig.4. The $conf$ results of each face resulting from commercial FR systems are pasted at the bottom. The red numbers before the slash are on Face++ and the blue number after the slash are from Microsoft.
\label{fig:mkup_present}}
\end{figure*}

\noindent{\bf Datasets.} Two public datasets are utilized in the following experiments: 
1) LFW \cite{huang2008lfw} contains 13233/5749 web-collected images/subjects with 6000 comparisons of same/different identities, most of them are low-quality. We randomly select 1000 different identity pairs to evaluate the attack performance. 2) \cite{gu2019ladn} released a high-quality makeup face database, including 333 frontal before-makeup faces and 302 after-makeup faces. By doing makeup generation, we select 198 after-makeup images with eye-makeup.
In our digital attack setting, random 100 source before-makeup faces and 10 targets form 1000 comparisons for impersonation attacks. For both of the datasets, we report the average results across 1000 identity pairs.

\noindent{\bf Competitors.} 
We choose several general attack methods, \emph{i.e.}, FGSM \cite{goodfellow2014explaining}, PGD \cite{madry2017towards}, MI-FGSM \cite{dong2018boosting}, C\&W \cite{carlini2017towards}), and face attack works, \emph{i.e.}, Adv-Glasses \cite{sharif2016accessorize}, Adv-Hat \cite{komkov2019advhat}, and Adv-Patch \cite{brown2017adversarial}, Adv-Makeup with ensemble training (W.O. Meta in Tab.\ref{Tab:results on digital-environments}), as the competing approaches. 
For a fair comparison, when comparing with the general attack methods, we restrict the attacking region only to the eye surroundings by multiplying the same 0-1 mask as ours.

\noindent {\bf Implementation details.} 
Our framework is implemented via PyTorch. The architectures of encoder and decoder are based on LADN \cite{gu2019ladn} with a U-Net structure. 
We employ an initial learning rate as 0.001 with momentum 0.9 via SGD. 
To balance different losses effects in 
Eq.\ref{eq: metaopt}, we set $ \alpha_1, \alpha_2, \beta_1, \beta_2,\beta_3 $ to be $ 1, 1, 0.1, 0.1, 0.1 $, respectively.
We choose IR152, IRSE50, MobileFace \cite{deng2019arcface} and Facenet \cite{schroff2015facenet} as the victim FR models, $ 3 $ of them are selected for white-box training, and the remaining one is the black-box model. For all the comparison methods, we follow their official experimental settings. 
More experimental results please refer to the supplementary materials.

\begin{table*}
\begin{centering}
\setlength{\tabcolsep}{1.6mm}{
\hspace{-0.1in}\scalebox{1.0}{ %
\begin{tabular}{c}
\begin{tabular}{c|c|cccc|cccc}
\hline 
\multirow{2}{*}{Methods } & Dataset & \multicolumn{4}{c|}{LFW Datatset} & \multicolumn{4}{c}{Makeup Datatset}\tabularnewline
\cline{2-10}
& Taget Model & IR152 & IRSE50 & FaceNet & MobileFace & IR152 & IRSE50 & FaceNet & MobileFace\tabularnewline
\hline 
\multirow{4}{*}{Gradient-based} & FGSM & 2.51 & 5.54 & 1.75 & 5.27 & 7.32 & 2.13 & 9.44 & 32.1 \tabularnewline
& PGD & 4.73 & 16.2 & 2.25 & 14.62 & 13.41 & 40.50 & 9.87 & 51.68 \tabularnewline
& MI-FGSM & 4.75 & 16.21 & 2.12 & 14.30 & 13.43 & 40.92 & 9.73 & 52.36 \tabularnewline
& C\&W & 1.64 & 6.95 & 1.23 & 5.56 & 8.59 & 28.66 & 21.34 & 33.71 \tabularnewline
\hline 
\multirow{3}{*}{Patch-based} & Adv-Hat & 3.65 & 15.71 & 2.75 & 20.36 & 8.04 & 34.22 & 12.21 & 41.36 \tabularnewline
& Adv-Glasses & 2.77 & 12.15 & 3.26 & 12.36 & 8.05 & 28.91 & 14.33 & 34.35 \tabularnewline
& Adv-Patch & 1.23 & 9.35 & 4.65 & 10.32 & 5.93 & 24.09 & 14.46 & 31.11 \tabularnewline
\hline 
\multirow{2}{*}{Ours} & W.O. Meta & 5.23 & 15.91 & 5.06 & 19.55 & 21.43 & 56.17 & 32.61 & 61.62 \tabularnewline
 & Adv-Makeup & \textbf{7.59} & \textbf{17.16} & \textbf{5.98} & \textbf{22.03} & \textbf{23.25} & \textbf{59.06} & \textbf{33.17} & \textbf{63.74}  \tabularnewline
\hline 
\end{tabular}\tabularnewline
\end{tabular}}}
\par\end{centering}
\caption{ASR results of digital impersonation attack over makeup and LFW datasets. There are four models used for the experiments, \emph{i.e.}, IR152, IRSE50, MobileFace, and FaceNet. For each column, we regard the written model as the target black-box model, and the rest three models are used for training.
The results are all from black-box attack setting. 
\label{Tab:results on digital-environments}}
\end{table*}

\noindent{\bf Attacking protocols.} 
The \emph{attack success rate (ASR)} ~\cite{deb2019advfaces,zhong2020towards} is reported as the evaluation metric,

\begin{center}
\vspace{-5mm}
\begin{equation}
\emph{ASR}=\frac{\sum_i^N 1_{\tau} ( cos[F(I_t^i), F(\hat{I_s^i})] > \tau)}{N} \times 100\%
\end{equation}
\par\end{center}
where $ 1_{\tau} $ denotes the indicator function. We mainly consider impersonating attack, therefore the proportion of comparisons with similarity larger than $ \tau $ will be obtained as ASR. The value of $ \tau $ will be set as the threshold at 0.01 FAR (False Acceptance Rate) for each victim FR model as most face recognition works do, \emph{i.e.}, IR152 ($ 0.167 $), IRSE50 ($ 0.241 $), MobileFace ($ 0.302 $) and Facenet ($ 0.409 $). 

Meanwhile, to further explore the reliability of attacks, we choose two cross-platform commercial state-of-the-art online face recognition systems as black-box victim systems to attack, \emph{i.e.}, Face++ \cite{MEGVII} and Microsoft \cite{Microsoft}. Thus, we also report the \emph{confidence score (conf)}, which means the confidence between attack image and target one, returned by the online systems in visualization results.

\subsection{Digital-attack}
To prove the effectiveness of the proposed Adv-Makeup on digital images, we design the multi-model attack experiments to present the performance improvement beyond the traditional ensemble adversarial training applied by other attacks.

\noindent{\bf Comparison with other attack methods.}
As mentioned before, we select seven different attack methods to compare, where training and testing paradigms strictly follow the original setting. As shown in Tab.\ref{Tab:results on digital-environments}, each column represents the ASR results on the remaining black-box victim model for transferable testing. 
Comparing to other attacks, Adv-Makeup achieves the best transferable results on all victim models and significantly outperforms the competitors. 

\begin{figure}
\begin{centering}
\includegraphics[scale=0.28]{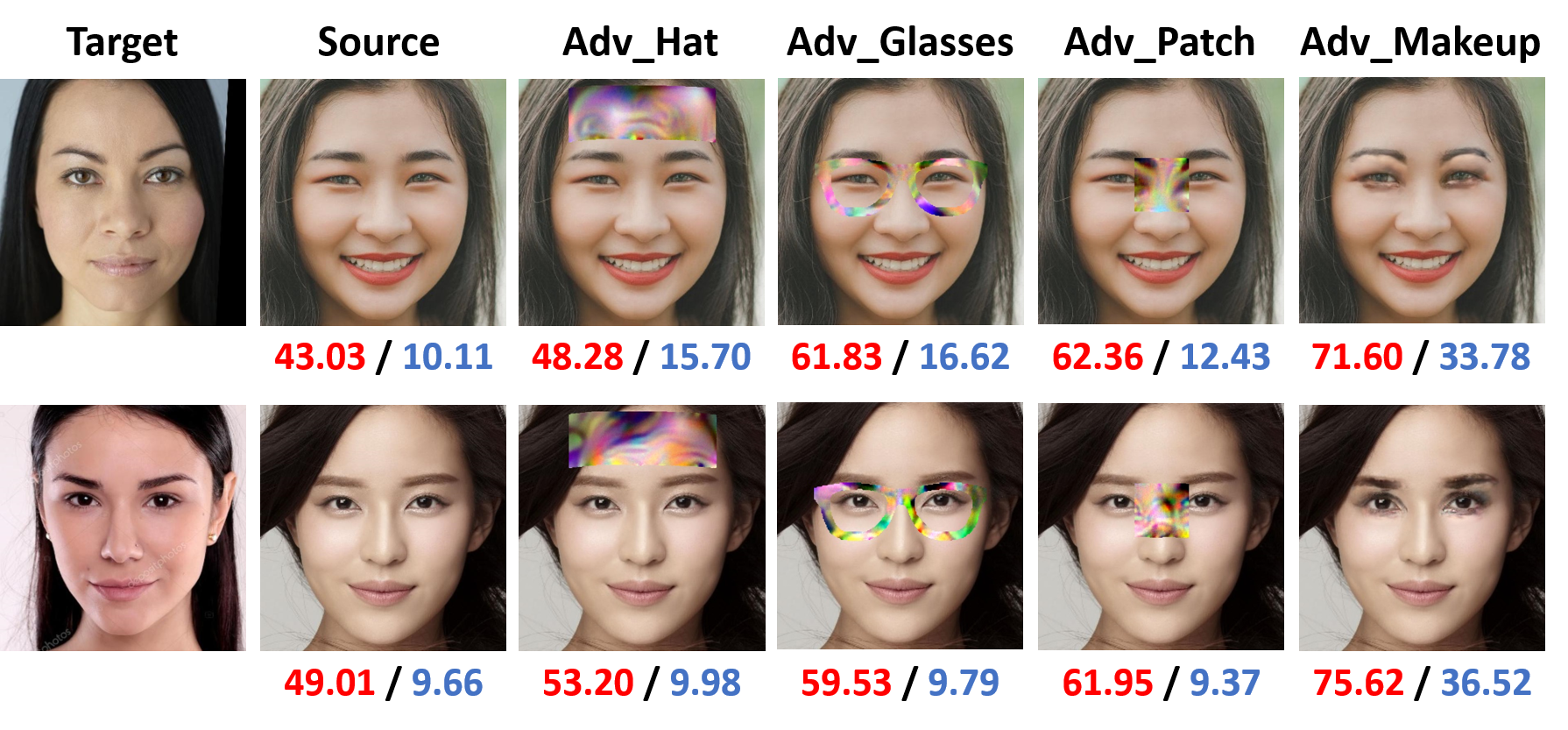}
\par\end{centering}
\caption{Visual comparison with other face attack methods. The generated adversarial faces attack two black-box commercial FR platform, and the $conf$ results of each attack are pasted at the bottom. The red numbers before the slash are the $conf$ results on Face++ and the blue numbers after the slash are from Microsoft.
\label{fig:digital_vis}}
\end{figure}

\noindent{\bf Quality of generated adversarial faces.} 
To exhibit the imperceptibility of Adv-Makeup, we show six adversarial faces generated by Adv-Makeup in Fig.\ref{fig:mkup_present}, two numbers below each face image are the $conf$s given by the commercial online FR systems. The adversarial faces and corresponding $conf$ results show that our method is robust to various interference factors such as illumination, pose, gender, and the intensity of eye makeup.
We also compare the visual results with Adv-Hat, Adv-Glasses, and Adv-patch. 
In Fig.\ref{fig:digital_vis}, the  examples show that the attacking makeup patches generated by Adv-Makeup own the most visually-indistinguishable appearance, while other forms need larger size of attacking area and look artificially and unnaturally to humans. Besides, we test the attacks on two commercial FR systems, and the comparison quantitative results again highlight the strength of Adv-Makeup in transferable black-box attack. 

\begin{figure}[t]
\begin{centering}
\includegraphics[scale=0.28]{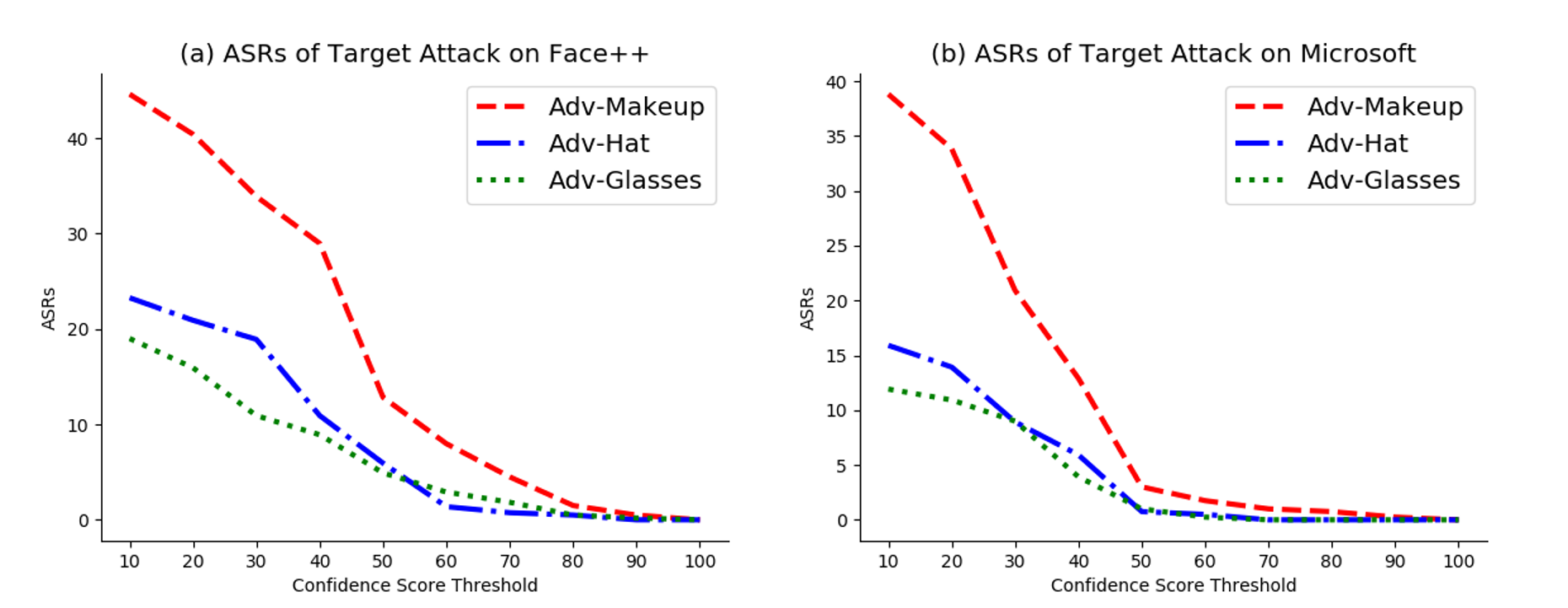} 
\par\end{centering}
\caption{The ASR comparison results of target attacks along with the confidence score threshold changes on Face++ and Microsoft.
\label{fig:ASRs_Threshold}}
\end{figure}

\subsection{Physical-attack}
Comparing to digital-attack, physical-attack has more practically real-world applications.
Thus, we utilize a simple tattoo paster technology to produce the Adv-Makeup attacks, and paste the  tattoo eye shadow over attacker orbital region.
Attacking two popular commercial FR platforms, we choose 20 identities, balanced in gender and age, as personators, and randomly select 20 targets from makeup dataset.

\begin{figure}
\begin{centering}
\includegraphics[scale=0.23]{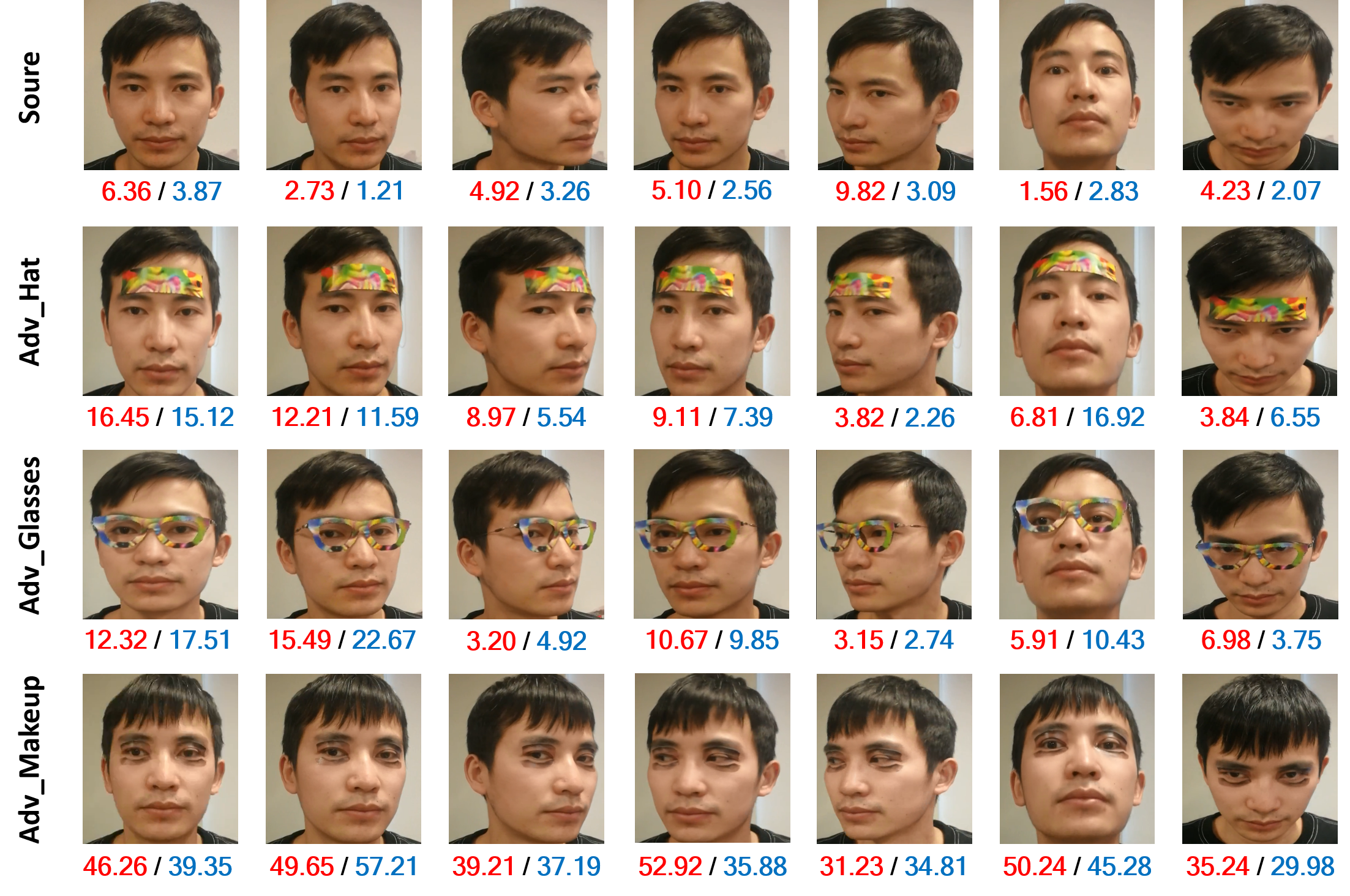} 
\par\end{centering}
\caption{Adversarial faces with {\bf pose variations} under {\bf physical-attack} against commercial FR platforms. The target identity is the same target female in the second row of Fig.\ref{fig:digital_vis}. The $conf$ results of each attack are pasted at the bottom, the red numbers before the slash are on Face++, and the blue after the slash are from Microsoft.
\label{fig:physical_vis_pose}}
\end{figure}

\noindent{\bf Adv-Makeup achieves the best physical attack results on the black-box commercial FR platforms.} In Fig.\ref{fig:ASRs_Threshold},
we show our average ASR results compared to Adv-Hat and Adv-Glasses according to different $conf$ threshold returned by the platforms, and some visualization results are illustrated in Fig.\ref{fig:physical_vis_pose}. It is clear that compared with current physical attacks, the attack strength of Adv-Makeup is substantially higher than the others, which demonstrates our method can generate feasible transferable adversarial faces in real applications.

\noindent{\bf Applying in real-world, the attacks of Adv-Makeup is the most imperceptible among the competitors.} As shown in Fig.\ref{fig:physical_vis_pose}, we show harder attacker pairs with different target gender, to compare with other attacks. It is obvious that our changing area is the smallest, and adversarial faces are the most natural and inconspicuous.

\noindent{\bf Adv-Makeup achieves better attack robustness to face pose changes.} Under real-world application scenes, most captured faces are not frontal, thus, the attack robustness to pose variations is crucial for physical attack. As shown in Fig. \ref{fig:physical_vis_pose}, our Adv-Makeup achieves the most stable attack ability under large pose changes.

\section{Conclusion}
To implement a more powerful and feasible attack for real-world face recognition applications, we propose a novel method Adv-Makeup which
can generate natural and seamless eye shadow to the orbital region, and achieve transferable attack under black-box models.
The visualization results show that our method can generate much more imperceptible adversarial examples than other competitors.
The experimental results on LFW and makeup datasets demonstrate the transferability of adversarial faces supervised by the task-driven fine-grained meta-learning adversarial attack strategy.
The case study targeting two commercial FR platforms further verify the feasibility and efficacy of Adv-Makeup.

\bibliographystyle{named}
\bibliography{ijcai21}

\end{document}